\documentclass[10pt,twocolumn,letterpaper]{article}

\usepackage[pagenumbers]{cvpr} 

\usepackage[dvipsnames]{xcolor}

\usepackage{multirow}
\usepackage{tabularx}
\usepackage{makecell}
\usepackage{array}
\usepackage[accsupp]{axessibility} 

\definecolor{cvprblue}{rgb}{0.21,0.49,0.74}
\usepackage[pagebackref,breaklinks,colorlinks,citecolor=cvprblue]{hyperref}

\def\paperID{6735} 
\def\confName{CVPR}
\def\confYear{2024}

\title{SFOD: Spiking Fusion Object Detector}

\author{
Yimeng Fan\textsuperscript{1}, 
Wei Zhang\textsuperscript{1,2,}\thanks{Corresponding author.}, 
Changsong Liu\textsuperscript{1}, 
Mingyang Li\textsuperscript{1}, 
Wenrui Lu\textsuperscript{1} \\
\textsuperscript{1}School of Microelectronics, Tianjin University, China\\
\textsuperscript{2}Tianjin Key Laboratory of Low-dimensional Electronic Materials and Advanced Instrumentation\\
\tt\small{\{yimengfan, tjuzhangwei, changsong, limingyang97, 3018232176\}@tju.edu.cn}
}

\begin{document}
\maketitle
\begin{abstract}
Event cameras, characterized by high temporal resolution, high dynamic range, low power consumption, and high pixel bandwidth, offer unique capabilities for object detection in specialized contexts. Despite these advantages, the inherent sparsity and asynchrony of event data pose challenges to existing object detection algorithms. Spiking Neural Networks (SNNs), inspired by the way the human brain codes and processes information, offer a potential solution to these difficulties. However, their performance in object detection using event cameras is limited in current implementations. In this paper, we propose the Spiking Fusion Object Detector (SFOD), a simple and efficient approach to SNN-based object detection. Specifically, we design a Spiking Fusion Module, achieving the first-time fusion of feature maps from different scales in SNNs applied to event cameras. Additionally, through integrating our analysis and experiments conducted during the pretraining of the backbone network on the NCAR dataset, we delve deeply into the impact of spiking decoding strategies and loss functions on model performance. Thereby, we establish state-of-the-art classification results based on SNNs, achieving 93.7\% accuracy on the NCAR dataset. Experimental results on the GEN1 detection dataset demonstrate that the SFOD achieves a state-of-the-art mAP of 32.1\%, outperforming existing SNN-based approaches. Our research not only underscores the potential of SNNs in object detection with event cameras but also propels the advancement of SNNs. Code is available at \url{https://github.com/yimeng-fan/SFOD}.
\end{abstract}    
\section{Introduction}
\label{sec:intro}

Event cameras are visual sensors that capture images in a novel manner. In contrast to conventional frame cameras that record complete images at a fixed rate, event cameras asynchronously collect changes in brightness at each pixel. Consequently, event cameras present noteworthy qualities, including high temporal resolution, high dynamic range, low power consumption, and high pixel bandwidth \cite{gallego2020event}. These attributes provide them with many advantages in object detection tasks, particularly in scenarios characterized by rapid motion or complex lighting conditions. However, the rapid sampling rate and the sparse, asynchronous format of event data present significant challenges to existing object detection algorithms. Consequently, it is crucial to address these challenges in the domain of event-based object detection research.

\begin{figure}[t]
  \centering
  \includegraphics[width=1.0\linewidth, height=0.5\linewidth]{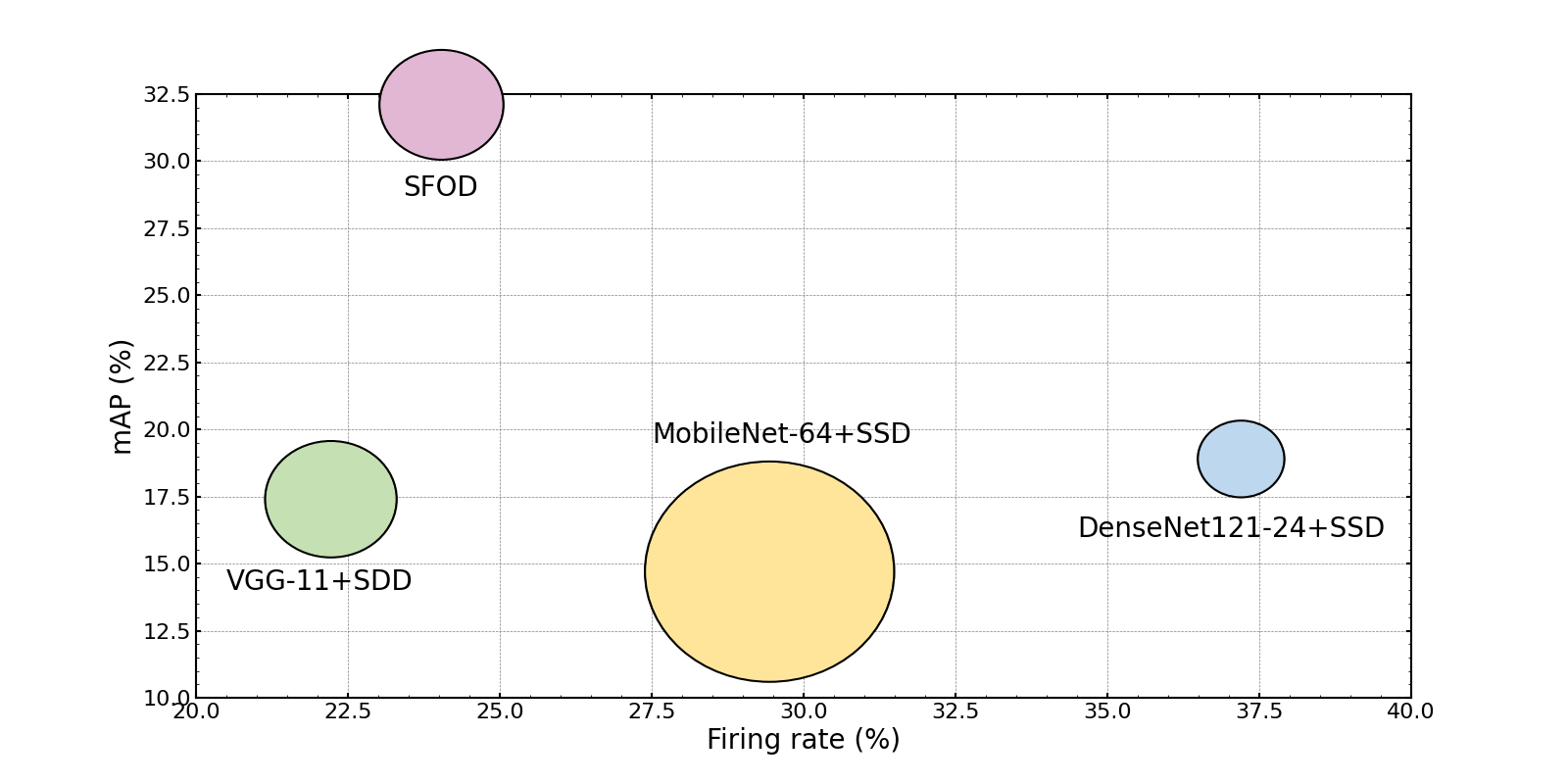}

   \caption{\textbf{Detection performance vs firing rate of our SFOD on the GEN1 dataset.}
   The areas of the circles correspond to the model size.}
   \label{fig:performance}
\end{figure}

Spiking Neural Networks (SNNs), recognized as the third generation of neural networks \cite{maass1997networks, roy2019towards}, are considered a promising solution. Unlike non-Spiking Neural Networks (non-SNNs), SNNs emulate the coding and processing of information in the human brain by utilizing spiking neurons as computational units \cite{maass1997networks}. This makes them inherently suited for processing event data. However, existing research on SNN-based object detection models applied to event cameras remains relatively limited \cite{cordone2022object}. Among them, the most crucial aspect that has not been thoroughly explored is the fusion of multi-scale feature maps. Such fusion is more important in SNNs than in non-SNNs, as it not only achieves a combination of deeper and shallower feature maps in the spatial domain but also enhances connections between features of different scales in the temporal domain. For instance, when recording a person walking with an event camera, the shallow layers of SNNs might initially detect only leg movements. However, as the person completes the movement, the deeper layers capture the full action. Multi-scale fusion in SNNs integrates these varying temporal perceptions from different layers, thereby significantly improving detection accuracy. In contrast, non-SNNs for RGB images only focus on spatial fusion and struggle with temporal data from event cameras, where RNNs can help but need more complexity and computational demands.

To tackle this problem, we propose the Spiking Fusion Module that realizes the fusion of multi-scale feature maps in SNNs applied to event cameras for the first time. This module is further combined with the Spiking DenseNet \cite{cordone2022object} and the SSD detection head \cite{liu2016ssd} to form the Spiking Fusion Object Detector (SFOD). In essence, our feature fusion strategy is as follows: multi-scale feature maps are extracted from the backbone network, which, after upsampling and concatenation, are fed into the Spiking Pyramid Extraction Submodule (SPES) to further refine the feature representations.

Furthermore, in SNNs, spike trains are used for the coding and processing of information. Therefore, decoding these spike trains by an effective strategy at the output layer is important for accurate inference. However, there is currently no corresponding research on spiking decoding for SNNs applied to event cameras. To investigate this, we perform a comprehensive analysis of different decoding strategies and conduct experiments with them during the pretraining phase of the backbone network. Concurrently, we study different classification loss functions to improve model performance. Our findings indicate that combining Spiking Rate Decoding with Mean Squared Error (MSE) loss function produces the best classification performance. As a result, we achieve state-of-the-art accuracy with SNNs on the NCAR dataset \cite{ncar}. In further experiments on the detection model, we evaluate the effect of different spiking decoding strategies and confirm that Spiking Rate Decoding can significantly improve the performance of the model.

The main contributions of this work can be summarized as follows:

(1)	We propose Spiking Fusion Module, which is the first to implement spiking feature fusion in SNNs for event cameras. It extracts and refines multi-scale feature maps from the backbone network in an SNN-friendly manner. This enhances the model's detection capabilities for targets of various scales. Furthermore, by integrating this module with Spiking DenseNet and SSD detection head, we design the SFOD, a simple and efficient SNN-based object detector.

(2)	For the first time in SNNs applied to event cameras, we conduct a thorough study of different spiking decoding strategies and classification loss functions to determine their impact on model performance. On the NCAR dataset, utilizing Spiking Rate Decoding paired with MSE loss, we achieve the state-of-the-art classification result based on SNNs, with an accuracy of 93.7\%.

(3)	On the GEN1 dataset \cite{gen1}, our SFOD achieves the state-of-the-art object detection performance of 32.1\% mAP for SNN-based models. Notably, compared to previous SNN-based detection models, SFOD not only demonstrates a significant enhancement in mAP but also maintains the model parameters and firing rate at a comparable level.

\begin{figure*}
  \centering
  \includegraphics[width=0.95\linewidth]{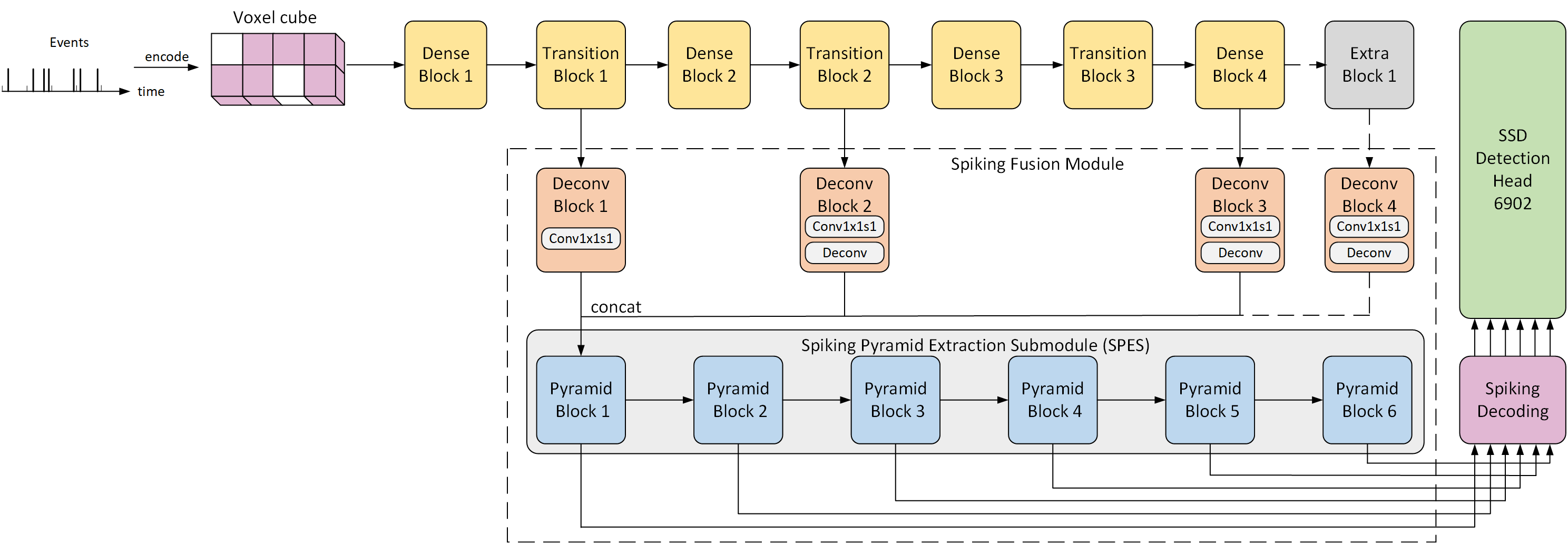}
  \caption{\textbf{The architecture of SFOD.} The Spiking Fusion Module is highlighted in the dotted area of the figure. In the fusion of layer four, Extra Block1 and Deconv Block4 are introduced and connected with the remainder of the network through the dotted lines.}
  \label{fig:SFOD_architecture}
\end{figure*}

\section{Related Work}
\label{sec:related}

\subsection{Spiking Neural Networks}\label{2.1}
SNNs can more closely mimic the spiking behavior of biological neurons than non-SNNs. To pursue this biomimicry, scholars have proposed various spiking neural models, including Hodgkin-Huxley (H-H) model \cite{H-H}, Izhikevich model \cite{izhikevich2003simple}, Leaky Integrate-and-Fire (LIF) model \cite{LIF}, and Parametric Leaky Integrate-and-Fire (PLIF) model \cite{PLIF}. Among these, the PLIF model, with its simplicity and capability to reduce the networks' sensitivity to initial conditions, has been widely adopted in current applications.

In SNNs, there are primarily three spiking decoding strategies: Spiking Count Decoding, Spiking Rate Decoding, and Membrane Potential Accumulation Decoding \cite{membrane}. Specifically, the Spiking Count Decoding counts spikes over a given duration, while the Spiking Rate Decoding divides this count by time \( T \), meanwhile the Membrane Potential Accumulation Decoding prohibits spike firing in the final layer and conveys information by accumulating membrane potentials.

SNNs primarily adopt two training strategies: ANNs-to-SNNs conversion and direct training. The ANNs-to-SNNs conversion uses the spiking rate to simulate the ReLU activation, enabling the transformation of trained ANNs into SNNs \cite{ann-snn1, ann-snn2}. Although this method has enabled the realization of powerful SNNs, such as Spiking YOLO proposed by \cite{spiking-yolo}, it requires thousands of time steps and is only suitable for static datasets rather than event-driven datasets. In contrast, the direct training approach employs surrogate gradient methods to optimize SNNs directly \cite{surrogate}, allowing SNNs to be trained on various datasets and achieve competitive performance within a few time steps. The success of this strategy has led to the widespread use of SNNs in visual tasks, including image classification \cite{classify_sew}, object detection \cite{cordone2022object}, and video reconstruction \cite{video}, etc. Consequently, we adopt the direct training strategy for our model in this work.

\subsection{Object Detection for Event Cameras}\label{2.2}
For object detection tasks with event cameras, one straightforward approach is to generate images frame-by-frame through temporal integration, thereby facilitating dense operations like convolution. Subsequently, traditional non-SNNs can be trained for object detection. This method has been extensively adopted in studies such as \cite{YOLOE, conventional2, conventional3, conventional4}. However, a significant drawback of this method is the loss of temporal information inherent in event data. To address this, algorithms like \cite{Matrix-LSTM, RED, RVT} have incorporated RNNs and Transformers, employing coding strategies that preserve temporal characteristics, thereby enhancing detection performance considerably. However, we believe that these techniques do not take full advantage of the sparsity inherent in event data as SNNs do. Although \cite{cordone2022object, su2023deep} pioneer the use of SNNs for object detection, the detection performance achieved remains suboptimal. Following these, we aim to further investigate and enhance SNN-based object detection algorithms applied to event cameras.

\section{Method}\label{sec:Method}
\subsection{Overview}\label{3.1}
The architecture of our proposed SFOD is illustrated in Figure \ref{fig:SFOD_architecture}. Initially, for the model to process sparse and asynchronous event data, we adopt the voxel cube \cite{cordone2022object} to code event data. Following this, we provide a brief explanation of the coding method. For event data \(\epsilon\) within the time interval \([t_a,t_b)\), the voxel cube can be expressed as:
\begin{equation}
E(\tau,c,x,y) = \sum_{e_k\in\epsilon} \delta \left ( \tau - \tau_{k}  \right ) \delta \left ( c - c_{k}  \right )\delta \left ( x - x_{k}, y - y_{k}  \right )
\end{equation}
\begin{equation}
\tau _{k} = \left \lfloor  \frac{t_{k} - t_{a}}{t_{a} - t_{b}} \cdot T \right \rfloor
\end{equation}
Herein, \(t_{k}\) indicates the timestamp of the event occurrence, \((x_{k},y_{k})\) denotes the pixel coordinates and \( T \) represents the number of time bins. To make more effective use of the temporal information contained in events, we further divide each time bin into \( n \) micro time bins. Combined with the event polarity \( P \in \{0, 1\}\), this results in a total of \( C \) channels, where \(C=2n\).

Then, to ensure efficient feature extraction, we employ Spiking DenseNet as the backbone network. This choice is motivated by its outstanding performance in prior work \cite{cordone2022object}. Subsequently, with the aim of extracting deeper feature maps from the backbone network, we adopt the Extra Block from \cite{cordone2022object}, which is composed of 1x1 and 3x3 convolutional layers. Based on this, the Spiking Fusion Module fuses and enhances the feature maps selected from the backbone network and the Extra Block. Finally, the processed feature maps are fed to the SSD detection head.

Notably, the SSD detection head consists solely of convolutions without any activation functions. Therefore, before feeding the features into it, we perform spiking decoding on these feature maps. Moreover, in order for the network to match the characteristics of SNNs, we replace all traditional activation functions with PLIF neurons.

In the following, we will elucidate the key design elements within the algorithm.

\subsection{Spiking Fusion Module}\label{3.2}
In the current SNNs applied to event camera object detection, there is a lack of studies on the fusion of multi-scale feature maps. To tackle this question, we propose the Spiking Fusion Module, a novel and efficient feature fusion module expressly designed for SNNs. With this method, we achieve the fusion within the spatial and temporal domains of the multi-scale feature maps. This improvement allows the model to better extract features and detect targets at various scales, resulting in a significant enhancement in its performance. The Spiking Fusion Module can be described as follows:
\begin{equation}\label{sfm}
X_{p} = \Phi _{p} \left ( \Phi _{f} \left ( \cup \left \{ \Phi_{i}\left ( X_{i} \right )  \right \} \right )   \right ) \quad i \in \mathcal{I},\  p \in \mathcal{P}
\end{equation}
In this formula, \(X_{i}\) where \(i \in \mathcal{I}\) represents the raw feature maps, while \(X_{p}\) with \(p \in \mathcal{P}\) denotes the newly generated feature maps post-fusion. The \(\Phi _{i}\) delineates the transformation function for the raw feature maps, \(\Phi _{f}\) stands for the feature fusion function and \(\Phi _{p}\) is the pyramid feature generation function. The \(\Phi _{i}\), \(\Phi _{f}\) and \(\Phi _{p}\) respectively correspond to the Deconv Block, concat and SPES in Figure \ref{fig:SFOD_architecture}.

\begin{figure}[t]
  \centering
  \begin{subfigure}{0.3\linewidth}
    \includegraphics[width=\linewidth]{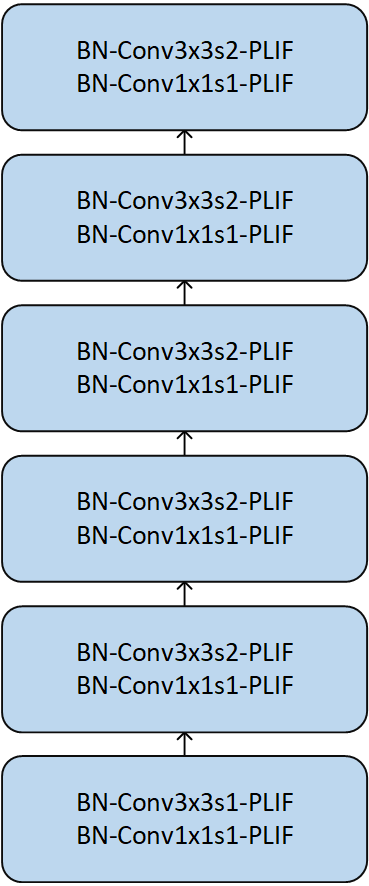}
    \caption{}
    \label{fig:sub1}
  \end{subfigure}%
  \hfill
  \begin{subfigure}{0.3\linewidth}
    \includegraphics[width=\linewidth]{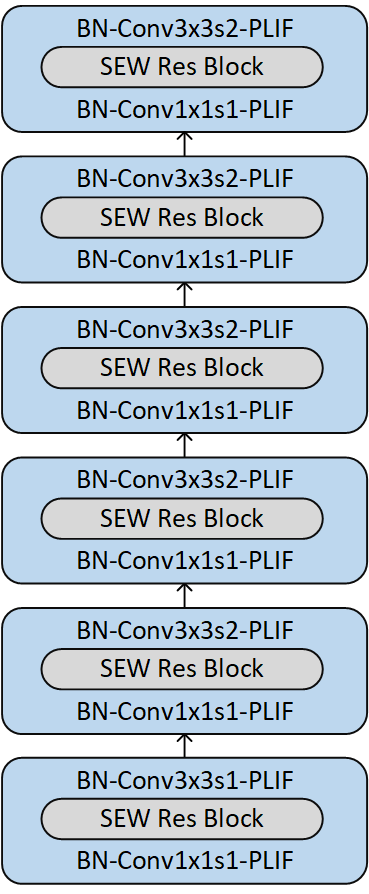}
    \caption{}
    \label{fig:sub2}
  \end{subfigure}%
  \hfill
  \begin{subfigure}{0.3\linewidth}
    \includegraphics[width=\linewidth]{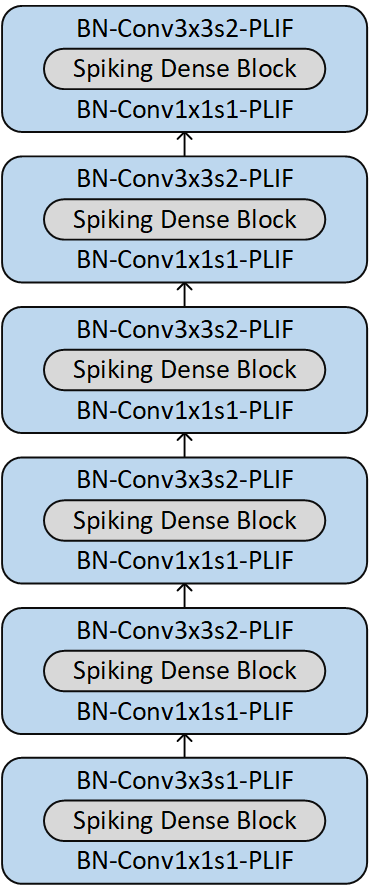}
    \caption{}
    \label{fig:sub3}
  \end{subfigure}
  \caption{\textbf{The architectures of SPES.} The blue block corresponds to the Pyramid Block in Figure \ref{fig:SFOD_architecture}.}
  \label{fig:pyramid}
\end{figure}

Next, we will discuss and analyze the design of \(\mathcal{I}\), \(\Phi _{i}\), \(\Phi _{f}\) and \(\Phi _{p}\) in the context of SNNs' characteristics.

\(\mathcal{I}\): In the \cite{cordone2022object}, the authors select six feature maps from Spiking DenseNet and its three appended Extra Blocks. These feature maps are then fed into the head network for object detection. The resolutions of these six feature maps are 30x38, 15x19, 7x9, 4x5, 2x3, and 1x2, respectively. We observe that although the deeper feature maps have a larger receptive field and can capture a broader context information, their spatial resolution is inadequate for effective fusion with the shallower layers. Consequently, we discard the last two feature maps (2x3 and 1x2) and conduct experiments to determine whether to retain the fourth feature map (4x5). As shown in Section \ref{4.3}, the performance of ignoring the fourth layer surpasses that of using it. Therefore, we opt to fuse the first three layers (30x38, 15x19, and 7x9).

\(\Phi _{f}\): In the non-spiking object detection models, there are primarily two methods for feature fusion. One involves fusing multi-scale feature maps by concatenation, such as ION \cite{ION}, HyperNet \cite{Hypernet}, and MFSSD \cite{small-FSSD}. The other connects different feature maps with element-sum, such as U-Net \cite{U-net} and FPN \cite{FPN}. However, in our perspective, employing the element-wise summation approach disrupts the binariness of SNNs, thereby increasing the computational complexity of the algorithm. Consequently, we adopt the method of concatenating multiple feature maps for fusion.

\(\Phi _{i}\): To employ the concatenation method for feature fusion, we upsample feature maps with resolutions smaller than 30x38 to ensure the same spatial dimensions across all feature maps. While bilinear interpolation is a prevalent upsampling technique in traditional frame-based image processing \cite{small-FSSD, deeplabv1}, this method involves numerous multiplication and division operations, which might impair the inherent binariness of SNNs. Consequently, we opt for transposed convolution (often referred to as deconvolution) \cite{deconv} for image upsampling. This approach not only preserves the characteristics of SNNs but also exhibits adaptability to spiking data. Furthermore, prior to the transposed convolution, we use a 1x1 convolution both to refine non-linear feature representations and to ensure the same channel dimensions across all feature maps, allowing the network to allocate equal attention to each feature map.

\(\Phi _{p}\): To regenerate and refine multi-scale feature maps, we propose the Spiking Pyramid Extraction Submodule (SPES). As illustrated in Figure \ref{fig:pyramid}, we present three variant SPES architectures. Architecture (a) is the basic version, in which the Pyramid Block is formed by 1x1 and 3x3 convolutions. As proposed by \cite{MS_CNN}, enhancing the head network of an object detection model can effectively boost its performance. However, in our model, the head network operates without spiking. Given this characteristic and the fact that the output of each Pyramid Block is fed into the head network, we aim to indirectly improve the head network's performance by enhancing SPES. Thus, we propose architectures (b) and (c) as enhanced versions of (a). Notably, with our network exceeding 100 layers and to avoid potential gradient vanishing from enhanced SPES, architectures (b) and (c) employ the Spike-Element-Wise Residual Block (SEW Res Block) \cite{classify_sew} and Spiking Dense Block \cite{cordone2022object} for enhancement, respectively. Finally, we adopt the SEW Res Block to enhance SPES for the efficient extraction of multi-scale feature maps, as supported by the experimental results and analysis in Section \ref{4.3}.

The Spiking Fusion Module can be summarized as follows: First, the Deconv Block conducts 1x1 convolution and transposed convolution upsampling on the first three feature maps. Subsequently, the processed feature maps with the same spatial dimensions are concatenated for feature fusion. Finally, the fused feature maps are fed into the SPES to regenerate multi-scale feature maps.

\subsection{Spiking Decoding and Loss Function}\label{3.3}
As discussed in Section \ref{2.1}, there are primarily three spiking decoding strategies: Spiking Count Decoding, Spiking Rate Decoding, and Membrane Potential Accumulation Decoding. We contend that Membrane Potential Accumulation Decoding, by omitting the step of neuronal spike firing, not only disrupts the inherent information processing mechanisms of SNNs but also diminishes their non-linear expression capabilities. Therefore, our research focuses predominantly on Spiking Count Decoding and Spiking Rate Decoding. According to our analysis, the distinction between them lies in the normalization of output in Spiking Rate Decoding. This leads to a more uniform decoding result. 

Moreover, when pretraining the backbone network for classification task, our analysis indicates that the Mean Squared Error (MSE) loss is more appropriate than the Cross-Entropy (CE) loss. The reasons are as follows:

First, for the discrete spike counts or frequencies decoded by the model, MSE can compute the loss and gradients directly. However, CE requires an extra softmax step to convert these discrete values into probability distributions, increasing the computational complexity. Second, the gradients of MSE directly represent the difference between decoded values and the labels, while the gradients of CE reflect the difference between post-softmax probability values and the labels. This characteristic of CE is suitable for non-SNNs, as the continuous floating-point outputs of these networks, when processed via softmax, can closely approximate the labels. However, SNNs produce limited discrete output values. This leads to optimization issues in the neurons when using CE loss, consequently reducing the model's generalization ability and increasing its firing rate. For example, when the model employs Spiking Rate Decoding for binary classification and assumes label of \([0, 1]\), consider two cases with decoded outputs of \([0.2, 0.8]\) and \([0.2, 1.0]\). Using CE, the results of the softmax computations are \([0.35, 0.65]\) and \([0.31, 0.69]\), respectively. Even though the decoded value for the negative class remains constant, change in probability values leads to reduced gradients for the negative class neuron. Meanwhile, the gradients for the positive class neuron, while decreased, still remain. In comparison, MSE derives the gradients directly from the decoded output, ensuring consistent gradients for the negative class neuron. When the decoded value for the positive class reaches \(1.0\), its gradients reduce to zero, halting further optimization.

\begin{equation}\label{MSE}
\text{MSE} = \frac{1}{N} \sum_{i=1}^N \sum_{j=1}^C (y_{ij} - a_{ij})^2
\end{equation}

\begin{equation}\label{CE}
\text{CE} = -\frac{1}{N} \sum_{i=1}^N \sum_{j=1}^C y_{ij} \log(z_{ij})
\end{equation}

The formulas for MSE and CE are described in Equations \ref{MSE} and \ref{CE}, respectively. In these equations, \(N\) is the sample size, \(C\) is the number of classes, while \(y_{ij}\), \(a_{ij}\) and \(z_{ij}\) represent the label, decoded value, and post-softmax probability value of the \(j^{th}\) class for the \(i^{th}\) sample, respectively.

Building on the above analysis, we conduct experiments on the NCAR dataset evaluating all combinations of decoding strategies and loss functions, as elaborated in Section \ref{4.2}. It's noteworthy that the combination of Spiking Rate Decoding and the MSE loss yields the best classification results. Additionally, in object detection task, experiments also show that Spiking Rate Decoding outperforms Spiking Count Decoding. Consequently, we opt for Spiking Rate Decoding as the decoding strategy in SFOD.

\section{Experiment}
In this Section, we first investigate the effects of different spiking decoding strategies and loss functions, and then pretrain the backbone networks, both on the NCAR dataset. Subsequently, we conduct an ablation study on the GEN1 dataset for SFOD and compare the best-performing model with state-of-the-art methods.

\begin{table}
\centering
\begin{tabularx}{\linewidth}{@{\extracolsep{\fill}}lccccc@{}}
\toprule
Models & Dec. & \makecell{Loss\\Func.} & Params & Acc. & \makecell{Firing\\Rate}  \\
\midrule
\textbf{\makecell[l]{DenseNet\\ 121-16}} & \textbf{Rate} & \textbf{MSE} & \textbf{1.76M} & \textbf{0.937} & \textbf{14.70\%} \\
\makecell[l]{DenseNet\\121-16} & Count & MSE & 1.76M & 0.869 & 12.95\% \\
\makecell[l]{DenseNet\\121-16} & Rate  & CE  & 1.76M & 0.930 & 20.42\% \\
\makecell[l]{DenseNet\\121-16} & Count & CE  & 1.76M & 0.920 & 17.09\% \\
\bottomrule
\end{tabularx}
\caption{\textbf{Spiking decoding and loss functions.} Comparison of different spiking decoding strategies and loss functions on the NCAR dataset.}
\label{table1}
\end{table}

\subsection{Experiment Setup}\label{4.1}
\noindent\textbf{Datasets.} The NCAR dataset \cite{ncar} is a binary classification dataset, comprising 12,336 car samples and 11,693 background samples. Each sample has a duration of 100 ms and exhibits varying spatial dimensions. 

The GEN1 dataset \cite{gen1} is the first large-scale object detection dataset captured by event camera. It consists of over 39 hours of car videos recorded by the GEN1 camera. Bounding box labels for cars and pedestrians within the recordings are provided at frequencies between 1 to 4Hz, amassing over 255,000 labels in total. 

\vspace{\baselineskip}
\noindent\textbf{Implementation Details.}
We code all samples with voxel cube, using a T-value of 5 and a micro time bin of 2, as done in \cite{cordone2022object}. All models are trained using the AdamW optimizer \cite{adamw}, coupled with cosine learning rate scheduler \cite{cosine}. On the NCAR dataset, our models are trained for 30 epochs, with a batch size of 64, an initial learning rate of 5e-3, and a weight decay of 1e-2. On the GEN1 dataset, the training parameters include 50 epochs, a batch size of 16, an initial learning rate of 1e-3, and a weight decay of 1e-4. To increase diversity and balance the sample distribution across classes on the GEN1 dataset, we apply horizontal flipping as a data augmentation strategy. 

\vspace{\baselineskip}
\noindent\textbf{Performance Metrics.} For classification tasks, accuracy is the primary evaluation metric. For object detection tasks, the main evaluation metrics are mAP@0.5:0.95 and mAP@0.5. Another crucial metric for evaluating SNNs is the firing rate. It is defined as the average ratio of the number of neuron spikes to the total number of neurons across all time steps, representing the level of neuronal activity. On certain specialized hardware, computations occur only when spikes are emitted. As a result, SNNs with a lower firing rate can notably reduce power consumption.

\begin{table}
\centering
\begin{tabularx}{\linewidth}{@{\extracolsep{\fill}}lccc@{}}
\toprule
Models         & Params     & Acc.     & Firing Rate \\
\midrule
\textbf{DenseNet121-16} & \textbf{1.76M} & \textbf{0.937} & \textbf{14.70\%} \\
DenseNet121-24 & 3.93M      & 0.928    & 15.90\%  \\
DenseNet121-32 & 6.95M      & 0.923    & 24.87\%  \\
DenseNet169-16 & 3.16M      & 0.921    & 15.34\%  \\
DenseNet169-24 & 7.05M      & 0.923    & 19.70\%  \\
DenseNet169-32 & 12.48M     & 0.894    & 27.28\%  \\ 
\bottomrule
\end{tabularx}
\caption{\textbf{Spiking DenseNets architectures.} Classification performance of different Spiking DenseNets architectures on the NCAR dataset}
\label{table2}
\end{table}

\subsection{Analysis and Pretraining on NCAR Dataset}\label{4.2}
In this section, we first explore the impact of different spiking decoding strategies and loss functions on model performance. Then, using the optimal combination, we train Spiking DenseNets of various depths and growth rates to study their structural influence. Based on the results, we choose the best-performing models as the backbone networks for our detection models.

\vspace{\baselineskip}
\noindent\textbf{Analysis of Spiking Decoding and Loss Function.} According to the experimental results in Table \ref{table1}, the model using Spiking Rate Decoding has better accuracy than the model using Spiking Count Decoding at the same level of firing rate, regardless of the loss function used. Specifically, when utilizing the MSE loss, the model with Spiking Count Decoding exhibits an approximate 7\% decline in accuracy compared to that using Spiking Rate Decoding. Table \ref{table4} presents the results for object detection models, and rows 2 and 3 further confirm this conclusion. We believe that the primary reason for this difference lies in the consistent setting of the prediction range between  \([0, 1]\) for both classification and detection tasks. This makes Spiking Rate Decoding more suitable than Spiking Count Decoding, which has an output range of \([0, T]\) that does not match the prediction range. This mismatch could impact the model's learning efficiency, as it necessitates adjustments within a wider output range. Conversely, the output range of Spiking Rate Decoding is normalized to align with the prediction values, thereby more effectively reflecting prediction errors and enhancing the model's learning ability.

Furthermore, the MSE loss outperforms the CE loss when Spiking Rate Decoding is used, both in terms of accuracy and firing rate. This further substantiates the viewpoint we present in Section \ref{3.3}. However, when using Spiking Count Decoding, the MSE does not perform as well as CE. We believe that the transformation of decoded values into probability distributions during the softmax step of CE computation serves as a normalization, which reduces the effects of the un-normalized Spiking Count Decoding. On the other hand, the MSE is extremely sensitive to the deviation between the predicted values and the labels, so using the un-normalized Spiking Count Decoding could lead to a higher accumulation of errors, ultimately affecting the overall performance of the model.

\begin{table}
\centering
\begin{tabularx}{\linewidth}{@{\extracolsep{\fill}}lccc@{}}
\toprule
Methods                        & Networks      & Acc.           & Firing Rate \\
\midrule
HATS \cite{ncar}                & N/A           & 0.902          & - \\
HybridSNN \cite{Hybrid_snn_ann} & SNNs-CNNs     & 0.906          & - \\
YOLOE \cite{YOLOE}              & CNNs          & 0.927          & - \\
EvS-S \cite{EvS-S}              & GNNs          & 0.931          & - \\
\textbf{Asynet \cite{Asynet}}   & \textbf{CNNs} & \textbf{0.944} & \textbf{-}\\ 
\midrule
HybridSNN \cite{Hybrid_snn_ann} & SNNs          & 0.770          & -  \\
Gabor-SNN \cite{ncar}           & SNNs          & 0.789          & -  \\
SqueezeNet 1.1 \cite{cordone2022object}         & SNNs          & 0.846          & 25.13\%  \\
MobileNet-64 \cite{cordone2022object}           & SNNs          & 0.917          & 17.14\%  \\
DenseNet169-16 \cite{cordone2022object}         & SNNs          & 0.904          & 33.59\%  \\
\textbf{VGG-11 \cite{cordone2022object}}        & \textbf{SNNs} & \textbf{0.924} & \textbf{14.69\%}  \\ 
\midrule
\textbf{DenseNet121-16}        & \textbf{SNNs} & \textbf{0.937} & \textbf{14.70\%}  \\ 
\bottomrule
\end{tabularx}
\caption{\textbf{Comparison with state-of-the-art models on the NCARS dataset.}}
\label{table3}
\end{table}

\begin{table*}
\centering
\begin{tabular*}{1.0\textwidth}{@{\extracolsep{\fill}}lccccccccc@{}}
\toprule
\multirow{2}{*}{Models} & \multicolumn{2}{c}{Dec.} & \multicolumn{3}{c}{Fusion Layers} & \multirow{2}{*}{Params} & \multirow{2}{*}{mAP@0.5:0.95} & \multirow{2}{*}{mAP@0.5} & \multirow{2}{*}{Firing Rate} \\
\cmidrule(r){2-3} \cmidrule(l){4-6}
                        & Rate       & Count       & None         & 3         & 4         &                             &                               &                          &                           \\
\midrule
DenseNet121-16-SSD      & \checkmark &             & \checkmark   &           &            & 5.0M                        & 0.262                         & 0.517                    & 21.01\%                   \\
DenseNet121\_24-SSD     &            & \checkmark  & \checkmark   &           &            & 8.2M                        & 0.235                         & 0.445                    & 22.02\%                   \\
DenseNet121-24-SSD      & \checkmark &             & \checkmark   &           &            & 8.2M                        & 0.288                         & 0.553                    & 22.29\%                   \\
DenseNet169-16-SSD      & \checkmark &             & \checkmark   &           &            & 7.7M                        & 0.257                         & 0.507                    & 22.82\%                   \\
SFOD-B                  & \checkmark &             &              &           & \checkmark & 15.0M                      & 0.294                         & 0.570                   & 21.13\%                   \\
SFOD-B                  & \checkmark &             &              & \checkmark &           & 9.9M                        & 0.299                         & 0.575                   & 24.41\%                   \\
SFOD-D                  & \checkmark &             &              & \checkmark &           & 11.3M                       & 0.286                        & 0.558                    & 26.37\%                   \\
\textbf{SFOD-R} & \checkmark & & & \checkmark & & \textbf{11.9M} & \textbf{0.321} & \textbf{0.593}  & \textbf{24.04\%} \\
\bottomrule
\end{tabular*}
\caption{\textbf{Results of the ablation study on the GEN1 dataset.}}
\label{table4}
\end{table*}

\begin{table*}
\centering
\begin{tabular*}{1.0\textwidth}{@{\extracolsep{\fill}}lccccccc@{}}
\toprule
Method                     & Networks         & Detection Head   & Params & \makecell{mAP\\@0.5:0.95} & \makecell{Firing\\Rate}  & \makecell{Time\\(ms)} & \makecell{Energy\\(mJ)} \\
\midrule
Asynet \cite{Asynet}        & Sparse CNNs      & YOLOv1 \cite{yolov1}   & 11.4M  & 0.145        & - & - & $>$ 4.83 \\
AEGNN \cite{AEGNN}          & GNNs             & YOLOv1           & 20.0M  & 0.163        & - & - & - \\
Inception+SSD \cite{conventional3} & CNNs      & SSD \cite{liu2016ssd}& -   & 0.301          & -  & 19.4 & - \\
MatrixLSTM \cite{Matrix-LSTM}      & RNNs+CNNs & YOLOv3 \cite{yolov3}&61.5M & 0.310        & - & - & - \\
RED \cite{RED}              & RNNs+CNNs        & SSD              & 24.1M  & 0.400        & - & 16.7 & $>$ 24.08 \\
\textbf{RVT \cite{RVT}} & \textbf{Transformer+RNNs} & \textbf{YOLOX \cite{yoilox}} & \textbf{18.5M} & \textbf{0.472} & \textbf{-} & \textbf{10.2} & \textbf{-} \\
\midrule
MobileNet-64+SSD \cite{cordone2022object}   & SNNs             & SSD              & 24.3M  & 0.147        & 29.44\% & 1.7\textsuperscript{\dag} & 5.76 \\
VGG-11+SDD \cite{cordone2022object}         & SNNs             & SSD              & 12.6M  & 0.174        & 22.22\% & 4.4\textsuperscript{\dag} & 11.06 \\
DenseNet121-24+SSD \cite{cordone2022object} & SNNs            & SSD              &8.2M    & 0.189        & 37.20\%  & 4.1\textsuperscript{\dag} & 3.89 \\
\textbf{EMS-YOLO \cite{su2023deep}} & \textbf{SNNs} & \textbf{YOLOv3} & \textbf{14.4M} & \textbf{0.310}   & \textbf{17.80\%} & \textbf{-} & \textbf{-} \\
\midrule
\textbf{SFOD}              & \textbf{SNNs}    & \textbf{SSD}     & \textbf{11.9M}  & \textbf{0.321}  & \textbf{24.04\%}  & \textbf{6.7} & \textbf{7.26} \\
\bottomrule
\end{tabular*}
\caption{\textbf{Comparison with state-of-the-art models on the GEN1 dataset.} A \textsuperscript{\dag} indicates that the runtime is not directly available and is estimated in the local environment. The method for calculating energy consumption can be found in the supplementary material.}
\label{table5}
\end{table*}

\vspace{\baselineskip}
\noindent\textbf{Pretraining on NCAR Dataset.} Based on the experimental results above, we further investigate the impact of different architectures of Spiking DenseNets on performance. As shown in Table \ref{table2}, it can be observed that when the growth rate is fixed, an increase in the model depth leads to a rise in firing rate and a decline in accuracy. Similarly, with a fixed model depth, an increase in the growth rate raises the firing rate, while only slightly reducing the accuracy. Notably, for models with a depth of 169, the one with a growth rate of 24 outperforms the one with a growth rate of 16 in accuracy. Taking into account the trade-off between accuracy and firing rate, we select DenseNet121-16, DenseNet121-24, and DenseNet169-16 as the backbone networks for the detection models, thereby delving deeper into the influence of model architecture on detection performance.

Table \ref{table3} presents a comparison between our best-performing model and other state-of-the-art methods on the NCAR dataset. The results indicate that our model not only outperforms other SNN-based methods but also surpasses the majority of methods based on non-SNNs in terms of accuracy. The accuracy gap between our model and the best non-spiking model is only 0.7\%.

\subsection{Ablation Study on GEN1 Dataset}\label{4.3}
In this section, we first study the performance differences across object detection models using various backbone networks. Based on this, we select the best backbone and further analyze the impact of different fusion layers. Finally, we compare the performance of various SPES variants. We name the models using the basic, Spiking Dense Block-enhanced, and SEW Res Block-enhanced SPESs as SFOD-B, SFOD-D, and SFOD-R, respectively.

\begin{figure*}
  \centering
  \includegraphics[width=0.98\linewidth]{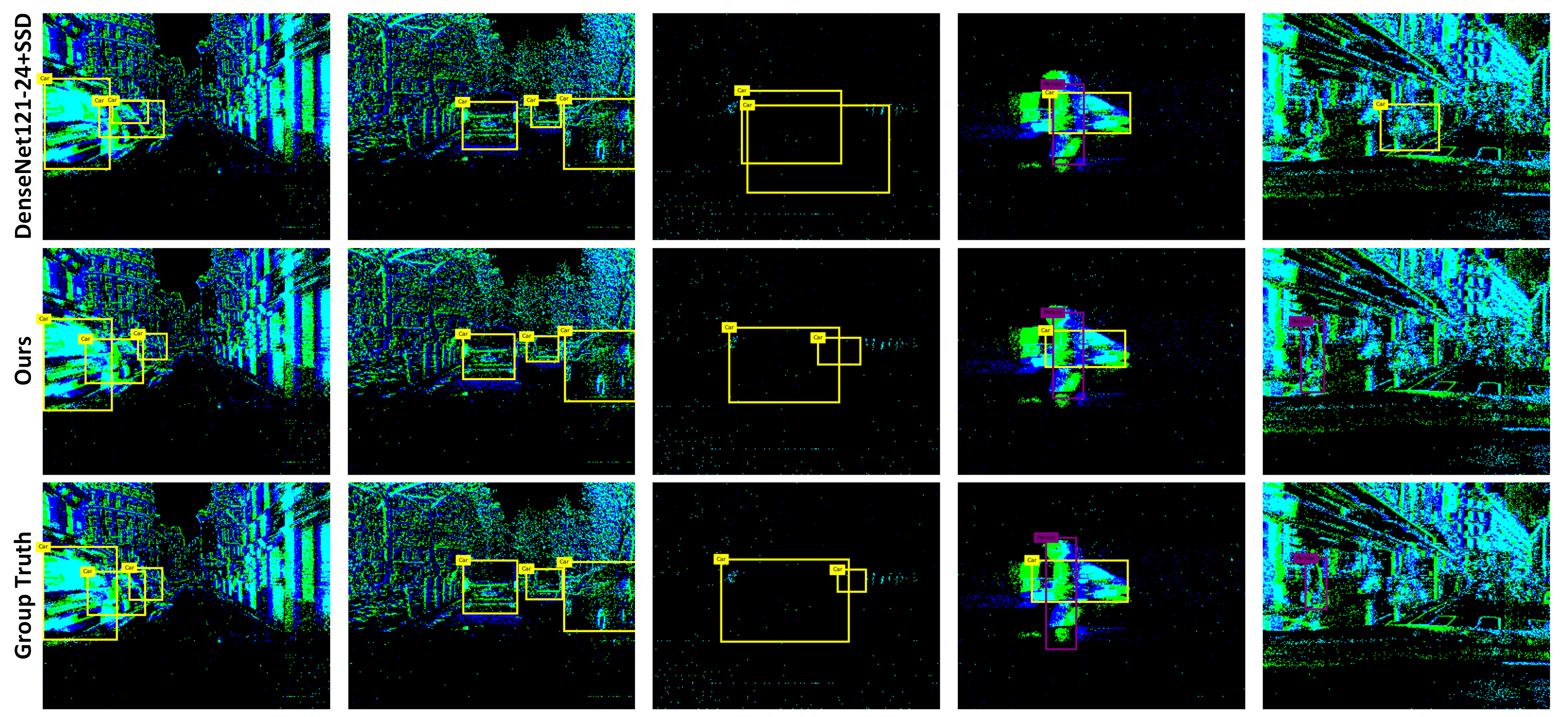}
  \caption{\textbf{Inference results of the model on the GEN1 dataset.} The figure illustrates the detection capabilities of the models across specific scenarios:  The first column demonstrates detection of overlapping cars; the second showcases non-overlapping detection; the third presents detection in sparse data contexts; the fourth reveals performance in multi-category scenes; and the fifth focuses on individual person target detection.}
  \label{fig:inference}
\end{figure*}

\vspace{\baselineskip}
\noindent\textbf{Different Backbone Network Architectures.} In Table \ref{table4}, specifically in rows 1, 3, and 4, we examine the effect of different backbone network architectures on model performance. Importantly, we don't incorporate the proposed Spiking Fusion Module into the detection models for this study. This decision not only leaves our experimental objectives unaffected but also provides a baseline for further experiments. The results indicate that when employing a consistent growth rate, an increase in network depth leads to a decline in mAP. However, at the same depth, a higher growth rate yields improved mAP. Furthermore, based on our observations, variations in network architectures appear to have no discernible impact on the firing rate. Therefore, we utilize DenseNet121-24 as the backbone network.

\vspace{\baselineskip}
\noindent\textbf{The Range of Fusion Layers.} As demonstrated in rows 5 and 6 of Table \ref{table4}, although fusing the fourth layer of feature maps effectively reduces firing rate, it does not lead to an improvement in mAP. In light of comprehensive consideration, we prioritize the improvement of mAP over the reduction of firing rate. Therefore, we conclude that a strategy of fusing three layers is superior to fusing four layers.

\vspace{\baselineskip}
\noindent\textbf{Comparison of Different SPESs.} Rows 6, 7, and 8 of Table \ref{table4} present a comparison of various SPES variants. The results reveal that, compared to the SFOD-B, the SFOD-D worsens both the firing rate and the mAP, while the SFOD-R not only shows a slight decrease in firing rate but also significantly improves the mAP by 2.2 points. This suggests that the identity mapping introduced by SEW Res Block can notably elevate model performance, whereas the multi-feature map connection mechanism introduced by Spiking Dense Block results in performance decline. Furthermore, while the integration of the SEW Res Block brings in certain non-spiking computations, these operations, which are limited to the six layers of 3x3 convolutions within SPES, rarely occur in our experiments. Therefore, we think these extra computations are acceptable compared to the significant improvements they bring to SPES.

\subsection{Benchmark Comparisons}\label{4.4}
In Table \ref{table5}, we present a comparison of our model with other state-of-the-art approaches on the GEN1 dataset. Remarkably, our model achieves a state-of-the-art mAP of 32.1\% at the same level of firing rate and parameters compared to other SNN-based methods. This performance nearly doubles that reported in \cite{cordone2022object}. Additionally, our model surpasses the majority of methods based on non-SNNs. When compared to RED \cite{RED} and RVT \cite{RVT}, our model not only has fewer parameters but also demonstrates significant advantages in both energy consumption and computation speed. 

Figure \ref{fig:inference} presents the results of our model in comparison with DenseNet121-24+SSD \cite{cordone2022object} and the Ground Truth. The DenseNet121-24+SSD \cite{cordone2022object} is a reproduced version in the local environment to ensure a fair comparison. From the figure, it is evident that our model consistently outperforms the other model in various scenarios.
\section{Conclusion}
In this paper, we propose a simple and efficient Spiking Fusion Module. Through this novel approach, we not only establish the current state-of-the-art SNN-based event camera object detection model, SFOD, but also achieve an impressive mAP performance on the GEN1 dataset. Compared to the method reported in \cite{cordone2022object}, the performance of SFOD is nearly double, representing a considerable advancement. Furthermore, during the pretraining phase of the backbone networks, we conduct an in-depth exploration of various combinations of spiking decoding strategies and loss functions. By adopting a combination of Spiking Rate Decoding and MSE, we establish the state-of-the-art SNN-based classification results on the NCAR dataset. More importantly, Spiking Rate Decoding has also significantly contributed to the enhancement in the performance of SFOD.

In the future, we believe that the performance of SFOD is expected to be further improved by adopting a more effective data augmentation strategy. It undeniably represents a promising research direction.\par
{
    \small
    \bibliographystyle{ieeenat_fullname}
    \bibliography{main}
}

\clearpage
\setcounter{page}{1}
\maketitlesupplementary

\setcounter{section}{0}
\renewcommand{\thesection}{\Alph{section}}

\section{Derivation of Classification Loss Function Gradients}
\label{A}
In this section, we present the derivation of gradients for the Mean Squared Error (MSE) and Cross-Entropy (CE) loss functions during backpropagation. To simplify, we assume the sample size of 1 and accordingly adjust the notation. The simplified formulas for MSE and CE are presented as Equations \ref{sim-MSE} and \ref{sim-CE}, respectively.

\setcounter{equation}{0}
\renewcommand{\theequation}{\Alph{equation}}
\begin{equation}\label{sim-MSE}
\text{MSE} = \sum_{j=1}^C (y_{j} - a_{j})^2
\end{equation}

\begin{equation}\label{sim-CE}
\text{CE} = - \sum_{j=1}^C y_{j} \log(z_{j})
\end{equation}

\subsection{Derivation of Mean Squared Error Loss Function Gradients}
The gradient derivation of the MSE loss function is detailed in Equation \ref{der_mse}. From this derivation, it can be concluded that the gradients of the MSE loss function are proportional to the difference between the decoded values and the labels.
\begin{equation}\label{der_mse}
\frac{\partial \mathrm{MSE}}{\partial a_j} = 2(a_{j} - y_{j}) 
\end{equation}

\subsection{Derivation of Cross-Entropy Loss Function Gradients}
The softmax function is shown in Equation \ref{softmax}:
\begin{equation}\label{softmax}
z_t= \frac{e^{a_t}}{\sum_{j=1}^{C} e^{a_j}}
\end{equation}

Define \(k\) as the index where \(y_k = 1\), the gradients of the CE loss function for \(j = k\) is given by Equation \ref{j=k}:

\begin{align}\label{j=k}
\frac{\partial \mathrm{CE}}{\partial a_j} &= \frac{\partial \mathrm{CE}}{\partial a_k} = \frac{\partial \mathrm{CE}}{\partial z_k} \frac{\partial z_k}{\partial a_k} \notag \\
&= -\frac{1}{z_k} \frac{(e^{a_k}) \sum_{j=1}^{C} e^{a_j} - e^{a_k}e^{a_k}}{\left( \sum_{j=1}^{C} e^{a_j} \right)^2} \notag \\
&= -\frac{1}{z_k} z_k(1 - z_k) \notag \\
&= z_k - 1
\end{align}

The gradients of the CE loss function for \(j \ne k\) is given by Equation \ref{jnek}:

\begin{align}\label{jnek}
\frac{\partial \mathrm{CE}}{\partial a_j} &= \frac{\partial \mathrm{CE}}{\partial z_k} \frac{\partial z_k}{\partial a_j} \notag \\
&= -\frac{1}{z_k} \frac{0 \cdot \sum_{j=1}^{C} e^{a_j} - e^{a_k}e^{a_j}}{\left( \sum_{j=1}^{C} e^{a_j} \right)^2} \notag \\
&= z_j
\end{align}

Equations \ref{j=k} and \ref{jnek} can be combined as follows:
\begin{equation}
\frac{\partial \mathrm{CE}}{\partial a_j} = z_j - y_j
\end{equation}

From this derivation, it can be concluded that the gradients of the CE loss function are proportional to the difference between the post-softmax probability values and the labels.

\section{Energy Consumption}
\label{B}
The low energy consumption advantage of SNNs mainly stems from performing accumulation calculation (AC) only when neurons fire. According to Section \ref{4.3}, although our model includes multiplication and addition (MAC) operations, such calculations are minimal and rarely occur in our experiments. Thus, MAC is not considered when evaluating the energy consumption of SNNs. In non-SNNs, network computations primarily rely on MAC operations. Although there are some AC operations, their limited number and significantly lower energy consumption compared to MAC allow us to disregard them for simplicity. In Table \ref{table5}, this is indicated by placing a greater than sign before the corresponding energy consumption values. Furthermore, in accordance with \cite{su2023deep}, we assume that data for different computations are realized as 32-bit floats in 45nm technology, with \(E_{\text{MAC}} = 4.6pJ\) and \(E_{\text{AC}} = 0.9pJ\). The formulas for calculating the energy consumption of SNNs and non-SNNs are presented as Equations \ref{energy_snn} and \ref{energy_non_snn}, respectively.

\begin{equation}\label{energy_snn}
E_{\text{SNNs}} = T \times fr \times E_{\text{AC}} \times N_{\text{AC}}
\end{equation}

\begin{equation}\label{energy_non_snn}
E_{\text{non-SNNs}} = T \times E_{\text{MAC}} \times N_{\text{MAC}}
\end{equation}

\section{More Implementation Details}
\label{C}

\renewcommand{\thefigure}{\Alph{figure}}
\setcounter{figure}{0}
\begin{figure*}\
  \centering
  \includegraphics[width=0.98\linewidth]{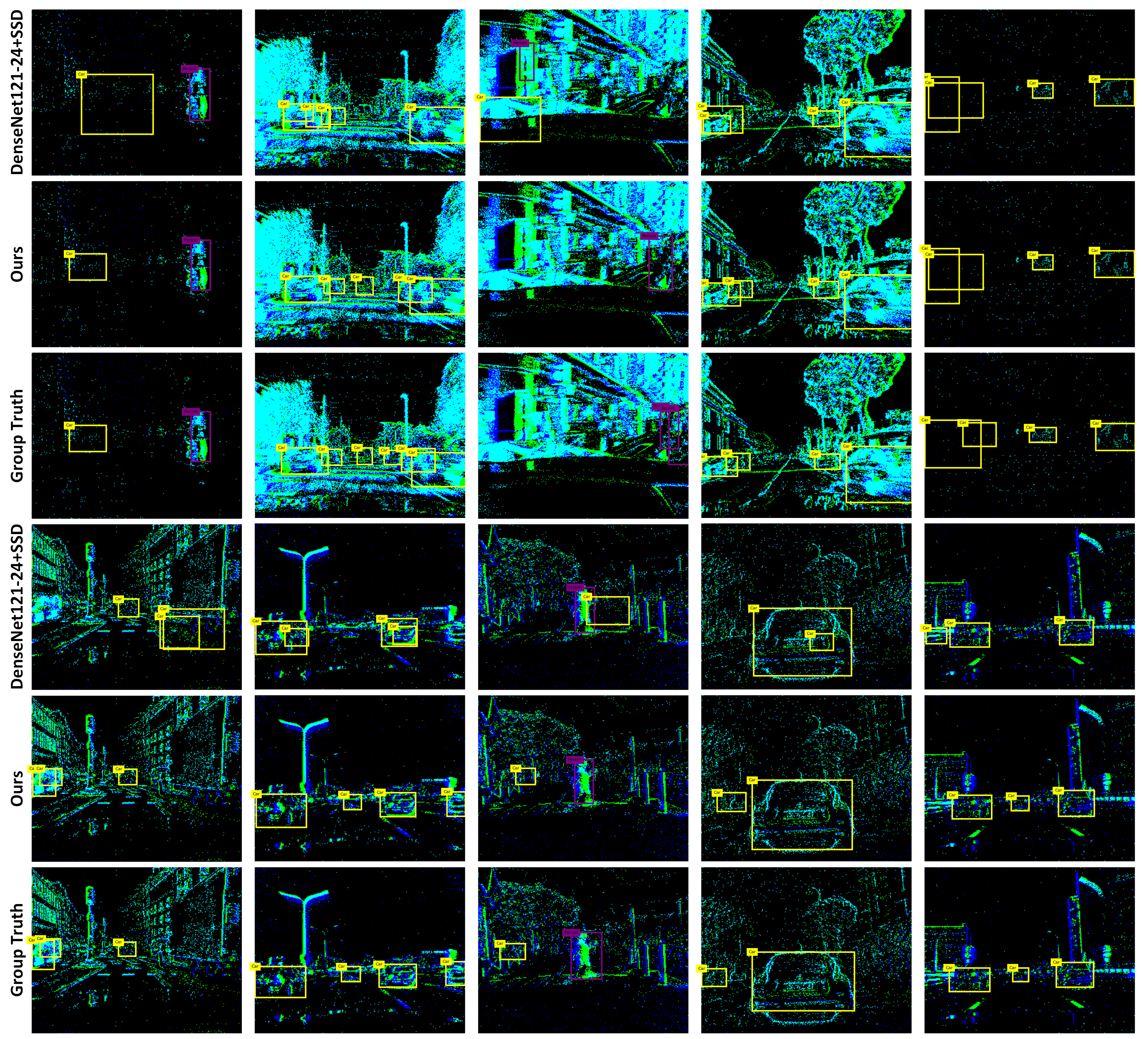}
  \caption{\textbf{More visual comparison results on the GEN1 dataset.}}
  \label{fig:more_vis}
\end{figure*}

On the NCAR dataset, to ensure consistent spatial dimensions for all samples, we employ the nearest-neighbor interpolation method to resize them to a resolution of 64x64. 

On the GEN1 dataset, given the 100 ms duration of NCAR dataset samples, accordingly, we extract the first 100 ms of each annotated box from this dataset as a sample. Furthermore, the samples in the GEN1 dataset have a consistent spatial resolution of 304x240. For the evaluation of the GEN1 dataset, following the criteria established in previous works \cite{li2022asynchronous, RED, RVT, cordone2022object}, we exclude bounding boxes with side lengths less than 10 pixels or diagonals shorter than 30 pixels.

Prior to each convolutional layer, we add a Batch Normalization (BN) layer \cite{batch}, recommended by \cite{cordone2022object} for better performance and faster convergence. All convolutional layers are initialized using the He Initialization method \cite{HeInitialization}, while biases in BN layers are set to 0, and weights are set to 1. The membrane time constant \(\tau\) for PLIF neurons \cite{PLIF} is initialized to 2, with the threshold set to 1. To mitigate gradient explosion, gradient norms are clipped at a maximum value of 1. We opt for Atan as the gradient surrogate function. All model training is executed on a single Nvidia A40 GPU. Furthermore, considering the class imbalance between the foreground and background in detection model training, we employ the Focal Loss \cite{focal} as the loss function. The formula for the Focal Loss can be described as Equation \ref{fl} (for simplicity, consider the binary classification case). In this formula, \(p_t\) as shown in Equation \ref{p_t} represents the predicted probability of a sample belonging to its true category. \(\alpha_t\) is a weighting factor for category balancing as shown in Equation \ref{alpha_t}. Lastly, \(\gamma\) is a hyper-parameter that is adjustable to fine-tune the behavior of the loss function.

\begin{equation}\label{fl}
FL = -\alpha_t (1 - p_t)^{\gamma} \log(p_t)
\end{equation}

\begin{equation}\label{p_t}
p_t = 
\begin{cases} 
p      & \text{if } y = 1, \\
1 - p  & \text{otherwise.}
\end{cases}
\end{equation}

\begin{equation}\label{alpha_t}
\alpha_t = 
\begin{cases} 
\alpha      & \text{if } y = 1, \\
1 - \alpha  & \text{otherwise.}
\end{cases}
\end{equation}

\section{More Visualization}
\label{D}
In this section, we present more visual comparisons of our model with DenseNet121-24+SSD \cite{cordone2022object} and the Ground Truth across a broader range of scenarios in Figure \ref{fig:more_vis}, further demonstrating the robust detection capabilities of SFOD in event camera object detection.

\end{document}